\documentclass[journal]{IEEEtran}

\ifCLASSINFOpdf

\else

\fi

\usepackage{graphicx, color, cite, epsfig, dsfont, amssymb, amsmath, amsthm, amsfonts, amsbsy, mathrsfs, mathtools, delarray}
\usepackage[colorlinks, linkcolor =blue, citecolor=blue, urlcolor=blue]{hyperref}
\usepackage[utf8]{inputenc}
\usepackage{subfig}
\usepackage{multirow}
\usepackage{algorithm}

\usepackage{algorithmic}
\usepackage{mathtools}
\usepackage{amsmath}
\usepackage{booktabs} 

\def\mclimits_#1{\limits_{\mathclap{#1}}}
\makeatother

\begin{document}

\title{Decentralized Multi-agent Reinforcement Learning based State-of-Charge Balancing Strategy for Distributed Energy Storage System }

\author{
Zheng Xiong,
Biao~Luo,~\IEEEmembership{Senior Member,~IEEE,} 
Bing-Chuan Wang, 
Xiaodong Xu,
Xiaodong Liu,
and 
Tingwen Huang,~\IEEEmembership{Fellow,~IEEE}}

\markboth{~}%
{Shell \MakeLowercase{\textit{et al.}}: Bare Demo of IEEEtran.cls for Journals}

\maketitle
\def\IEEEkeywordsname{Index Terms}

%
\begin{abstract}
This paper develops a {\it Dec}entralized {\it M}ulti-{\it A}gent {\it R}einforcement {\it L}earning (Dec-MARL) method to solve the SoC balancing problem in the distributed energy storage system (DESS). First, the SoC balancing problem is formulated into a finite Markov decision process with action constraints derived from demand balance, which can be solved by Dec-MARL. Specifically, the first-order average consensus algorithm is utilized to expand the observations of the DESS state in a fully-decentralized way, and the initial actions (i.e., output power) are decided by the agents (i.e., energy storage units) according to these observations. 
In order to get the final actions in the allowable range, a counterfactual demand balance algorithm is proposed to balance the total demand and the initial actions. 
Next, the agents execute the final actions and get local rewards from the environment, and the DESS steps into the next state. 
Finally, through the first-order average consensus algorithm, the agents get the average reward and the expended observation of the next state for later training. 
By the above procedure, Dec-MARL reveals outstanding performance in a fully-decentralized system without any expert experience or constructing any complicated model. Besides, it is flexible and can be extended to other decentralized multi-agent systems straightforwardly. Extensive simulations have validated the effectiveness and efficiency of Dec-MARL.

\end{abstract}


\IEEEpeerreviewmaketitle

\section{Introduction}
Microgrid is one of the most popular solutions for problems caused by renewable energy sources (RESs) integration and electrical loads increase. Microgrid can operate in parallel with the grid, as an autonomous power island or in transition between grid-connected mode and island mode of operation \cite{lidula2011microgrids,8825535,9349186}. A microgrid is formed by distributed loads, distributed RESs, and distributed energy storage system (DESS) \cite{TIE10006735}. Generally speaking, the DESS is critical to ensure that the microgrid works in a steady state.

As a significant component of the DESS, the energy storage units (ESUs) play a vital role in solving the primary problems faced by the microgrid. First, ESUs work as buffer zones for smoothing the RESs' fluctuation which is the key problem of RESs integration. Besides, ESUs can reduce users' power demands at critical times without changing the electricity consumption, which can solve the problems caused by exceeding electrical loads. Furthermore, ESUs can enable the microgrid to operate in the island mode \cite{morstyn2016control,TIEcarrasco2006power,ferahtia2022optimal}. 
However, since the output power of ESUs is allocated proportionally to their capacities, ESUs with lower state-of-charge (SoC) will run out of energy more quickly. Once an ESU is used up, it can no longer contribute its power capacity to the microgrid demand. Suppose the microgrid demand rises above the power capacity of the remaining ESUs, the microgrid frequency limits will be violated, and the remaining ESUs will be overloaded \cite{TIE9424403,morstyn2017multiwithoutRL,TIESOCsingle2022yang,TIESOCGuangzeShi2021ADS}, which brings danger to the stability of the system. Therefore, the SoC balancing problem, which can be regarded as a decision problem in the DESS, is critical to the safety of the microgrid \cite{jiang2019hierarchical,TII9099954,9720160}.

Reinforcement learning (RL) as an efficient method for decision problems has drawn lots of attention from researchers in recent years \cite{TIEMGRL2022demandresponse,liu2018distributed,sutton2018reinforcement}. Its applications in the field of microgrids have also been widely investigated \cite{TIESOCsingle2022yang,single2019intelligent,single2020deep,TIEsingle2022HEV,multi2020SamadiCTCE,li2019multiEM,zhu2022multiEM,CTDE2021multiEM,xuCTDE2020multiHEM,morstyn2017multiwithoutRL}. Without loss of generality, these studies can be classified into two categories: single-agent RL methods \cite{single2018HEV,single2019intelligent,TIEsingle2022HEV,TIESOCsingle2022yang,single2020deep} and multi-agent RL methods \cite{li2019multiEM,zhu2022multiEM,CTDE2021multiEM,xuCTDE2020multiHEM,multi2020SamadiCTCE}. In single-agent RL methods, all of the control variables are integrated into a single action vector.  Du and Li \cite{single2019intelligent} used a deep neural network-based RL method for multi-microgrid energy management to maximize the profit of selling power while protecting customer privacy. In \cite{single2020deep}, a deep reinforcement learning-based DESS controller was proposed, which can provide frequency response services to the power grid. Xiong $et \ al.$ \cite{single2018HEV} presented an RL-based real-time power management strategy to allocate power in a plug-in hybrid electric vehicle. However, single-agent RL methods will face the ``curse of dimensionality" when the scale of the considered system is large. 

Using multiple agents in RL is one way to alleviate the ``curse of dimensionality".
In \cite{CTDE2021multiEM}, a multi-agent deep Q-network was proposed to keep the benefits balance in solving the distributed energy management problem. 
Samadi $et \ al.$ \cite{multi2020SamadiCTCE} proposed a MARL-based decentralized energy management approach in a grid-connected microgrid. 
However, in most of the MARL methods, agents operate in the centralized training decentralized execution (CTDE) mode, which can only alleviate the ``curse of dimensionality" in the process of execution, and the process of training may still face this problem when there are a massive number of agents for large scale systems. Thus, the decentralized training decentralized execution (DTDE) mode without the need of any training center is more appropriate for a large-scale system like the DESS. To our knowledge, the fully-decentralized reinforcement learning method for the DESS is rarely investigated in the existing literature. The reasons are twofold. First, in a decentralized way, agents can only get incomplete observations of the DESS state, which is not enough for agents to decide effective actions. In addition, since agents only decide their own actions, it is difficult to satisfy the global action constraints in a decentralized multi-agent system.
\par Based on these observations, a fully-decentralized MARL-based energy management method (i.e., Dec-MARL) is proposed for solving the SoC balancing problem in the DESS.
The main contributions of this paper are summarized as follows:
\par $\bullet$ The SoC balancing problem in the time-ahead microgrid is formulated as a finite Markov decision process with discrete time steps, which is rarely investigated in the existing literature. 
\par $\bullet$ Agents' observations are expanded for more effective actions through the first-order average consensus algorithm in a fully-decentralized way. 
\par $\bullet$ A new counterfactual exploration method for the agents in the DESS is proposed, which can expand the exploration range while keeping the actions in the allowable range. As a result, the learning speed can be accelerated in the preliminary stage.

\par $\bullet$ Dec-MARL is developed for island microgrid by considering both SoC balancing and battery degradation with the constraints of energy management. 
Simulation results show the efficiency of our proposed method. 
\par The rest of the paper is organized as follows. Section \ref{Sec_2} describes some preliminary knowledge, and Section \ref{sec_3} formulates the problem as a mathematical problem. In Section \ref{Dec-MARL}, the Dec-MARL algorithm is presented. 
The simulation results are discussed in Section \ref{Simulations}, and conclusions are summarized in Section \ref{conclusion}.

\section{Preliminary} \label{Sec_2} 
In this section, the graph theory and the first-order average consensus algorithm applied in Dec-MARL are briefly introduced.

\subsection{Graph Theory} \label{graph theory}

Let $\mathcal{G}=(\mathcal{V},\mathcal{E},\mathcal{W})$ be a positively weighted undirected graph, where $\mathcal{V}$ is the set of nodes with $\mathcal{V}=\{v_i|i\in\{1,2,\dots,N\}\}$ and $\mathcal{N}$ is the set of ESUs in the DESS, 
$\mathcal{E}$ is the set of the edges between every two nodes with $\mathcal{E}\subseteq\mathcal{V}\times\mathcal{V}$, and $\mathcal{W}$ is the set of edge weights. An undirected edge between nodes $i$ and $j$ is denoted by a pair $(v_i,v_j)\in\mathcal{E}$, which means nodes $i$ and $j$ can communicate with each other. 
Let $(v_i,v_i)\in\mathcal{E}$ for any $v_i\in\mathcal{V}$ as each node has access to get its local information at any time. 
For simplicity, we define the neighbor set as a set of all the nodes that can get the information from node $i$, denoted by $\mathcal{N}_i=\{v_j|(v_i,v_j)\in\mathcal{E},\forall v_j\in\mathcal{V}\}$. 
If there is a connected path between every two nodes, then the graph $\mathcal{G}$ is called strongly connected. 

\subsection{The First-order Average Consensus Algorithm}\label{average consensus}
The first-order average consensus algorithm \cite{xiao2007distributed} to obtain the average of a vector is iterated as follows:
\begin{equation}\label{eq_2.1}
x_i[k+1]=x_i[k]+\sum_{j\in \mathcal{N}_i} w_{ij}(x_j[k]-x_i[k])
\end{equation}
where $x_{i(j)}[k]$ is the state of node $i(j)$ at the $k \text{th}$ iteration, and $ w_{ij}$ is the weight associated with the edge $\{i,j\}$. Since the graph is defined as an undirected graph, we have $w_{ij}=w_{ji} $. Setting $ w_{ij}=0 $ for $ j \notin \mathcal{N}_i$ and $w_{ii}=1-\sum_{j \in \mathcal{N}_i} w_{ij}$, the above iteration can be simplified as:
\begin{equation}\label{eq_2.2}
X[k+1]=WX[k]
\end{equation}
where $X[k]=[x_1[k],x_2[k],\dots,x_N[k]]^\mathsf{T} \in \mathcal{R}^N$ is the vector of $N$ nodes' states at the $k \text{th}$ iteration, and $W$ is the weight matrix.

If the matrix $W$ satisfies:\par
1) The sum of entries of any row vector or column vector is equal to 1.\par
2) The maximum eigenvalue is less than 1.\\
Then $x_i$ will converge to the average of initial states:
\begin{equation}\label{eq_2.4}
x_i^*=\frac{1}{N}\sum_{j=1}^Nx_j[0].
\end{equation}

\section{Problem Formulation}\label{sec_3}
\par
\begin{figure}[!t]
	\centering	\includegraphics[width=3.3in]{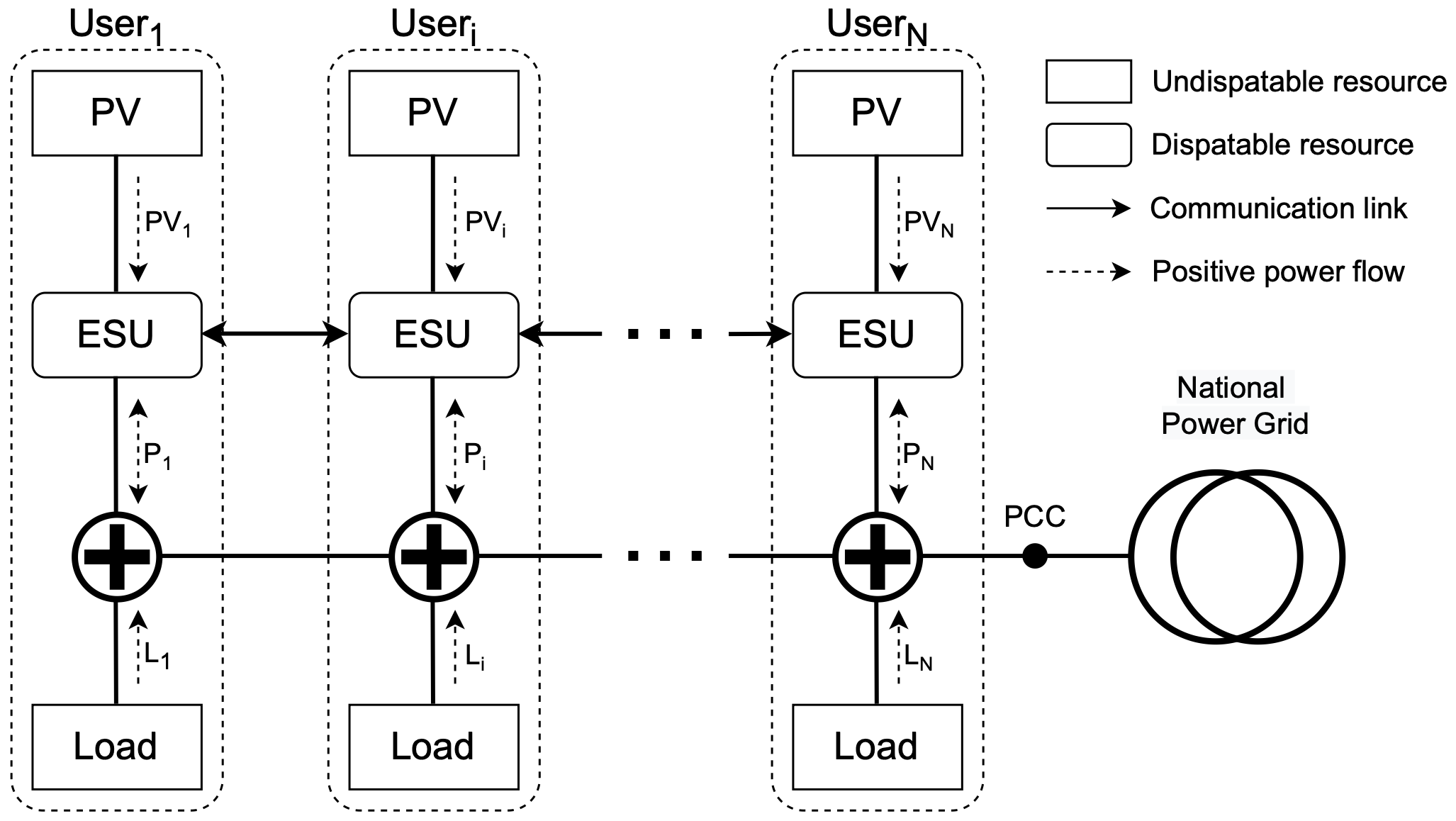}
	\caption{The simplified structure of the microgrid.} \label{fig1}
\vspace{-0.4cm}\end{figure}

 Fig. \ref{fig1} presents the simplified structure of the microgrid system. 
In a microgrid, there are three types of power components: loads from the users, RESs from distributed household photovoltaic (PV) panels, and the DESS consisting of household ESUs.
All of the components and the national power grid are connected through a point of common coupling (PCC).
The power flow and communication links are marked in Fig. \ref{fig1}. 

In this paper, the island mode, where the link between the microgrid and the national power grid is cut off, is considered. 
Thus, the DESS needs to satisfy the users' loads according to the RESs while balancing the SoC values of ESUs. 
The SoC balancing problem is then formulated into a mathematical problem with constraints.

\subsection{Energy Management Model}\label{sec_3.1}
The objective of energy management is to find an operating policy for allocating the output power of ESUs to satisfy the total power demand, which can be formulated as:
\begin{equation}\label{EMM}
	\sum_{i=1}^N P_{i,t}=\sum_{i=1}^N(L_{i,t}-PV_{i,t})
\end{equation}
where $P_{i,t}$ is the output power of the $i \text{th}$ ESU, $L_{i,t}$ denotes the $i \text{th}$ local user load, $PV_{i,t}$ denotes the $i \text{th}$ PV power, and $D_{i,t}=L_{i,t}-PV_{i,t}$ is defined as the local power demand at time step $t$.

\subsection{Energy Storage Unit Model}\label{sec_3.2}
Let $E_{i,t}$ denote the SoC of the $i$th ESU at the time step $t$. The simplified SoC model of the ESU can be given as follows:
\begin{equation}\label{ESUM}
	E_{i,t} =\left\{
	\begin{aligned}
	& E_{i,t}-\frac{P_{i,t}}{\eta_i C_i} & P_{i,t}>0\\
	& E_{i,t}+\frac{\eta_i P_{i,t}}{C_i}& P_{i,t}<0
	\end{aligned}
	\right.
	\end{equation}
where $C_i$ denotes the capacity of the $i$th ESU, and $\eta_i $ is the charging/discharging efficiency of the $i$th ESU. 
The output power of the $i$th ESU at time step $t$ must be in the allowable range $\underline{P_i} \le P_{i,t}\le \overline{P_i}$ decided by its physical properties, where $\overline{P_i}$ and $\underline{P_i}$ are the maximum and minimum output power of the $i$th ESU, respectively.

\par In addition to this, the output power of the $i$th ESU is also subjected to its current SoC:
\begin{equation}\label{eq_3.6}
	\frac{(E_{i,t}-E_i^{max})C_i}{\eta_i}\le P_{i,t}\le \eta_i(E_{i,t}-E_i^{min})C_i
\end{equation}
where $E_i^{min}$, $E_i^{max}$ are the allowed minimum, maximum SoCs of the $i$th ESU. 

\par According to the limitation discussed above, the bounds of the output power of the $i$th ESU at time step $t$ should be adjusted:
\begin{equation}\label{upper}
	\overline{P_{i,t}} = \min\{ \eta_i(E_{i,t}-E_i^{min})C_i,\overline{P_i}\}
\end{equation}
\begin{equation}\label{lower}
	\underline{P_{i,t}} = \max\{ \frac{(E_{i,t}-E_i^{max})C_i}{\eta_i}, \underline{P_i}\}.
\end{equation}
\par Besides, the inevitable degradation of ESU due to cumulative throughput during operation should be acknowledged. 
In this paper, the degradation model in \cite{antoniadou2020market} is used to describe the degradation caused by cumulative throughput:
\begin{equation}\label{battery degradation1}
	Q^{loss}_{i,t} = B_1 e^{B_2 I_c}	|P_{i,t}|\Delta T 
\end{equation}
where $Q^{loss}_{i,t}$ is the capacity loss of $i$th ESU at time step $t$, $ B_1$ and $ B_2$ are parameters obtained from experimental data, and $I_c$ is the average C-rate, which is a characteristic of the ESU \cite{antoniadou2020market}. 
The cost of degradation $C^{ct}_{i,t}$ can be written as:
\begin{equation}\label{battery degradation2}
	C^{ct}_{i,t} = \frac{C^B_i Q^{loss}_{i,t}}{100\% - \eta_B}
\end{equation}
where $C^B_i$ is the installation cost of the $i$th ESU, and $\eta_B$ is the end-of-life retained capacity percentage. 

\subsection{SoC Balancing Problem}

According to Sections \ref{sec_3.1} and \ref{sec_3.2}, the SoC balancing problem can be formulated as a mathematical problem with constraints
\begin{equation}\label{equ_10}	
	\begin{aligned}	
		\min\;&\sum_{t=0}^T\sum_{i=1}^N(\alpha(E_{i,t}-\overline{E_t})^2+\beta C^{ct}_{i,t})\\
&\text{s.t.} \quad E_{i,t} =\left\{
	\begin{aligned}
	& E_{i,t}-\frac{P_{i,t}}{\eta_i C_i} & P_{i,t}>0\\
	& E_{i,t}+\frac{\eta_i P_{i,t}}{C_i}& P_{i,t}<0
	\end{aligned}
	\right.\\
& \qquad \sum_{i=1}^N P_{i,t}=\sum_{i=1}^N(L_{i,t}-PV_{i,t})\\
& \qquad \underline{P_{i,t}} \le P_{i,t}\le \overline{P_{i,t}}\\
	\end{aligned}
\end{equation}
where $\overline{E_t}$ is the average SoC value of all ESUs, given by $\overline{E_t}=\frac{1}{N}\sum_{i=1}^N E_{i,t} $, $\alpha$ and $\beta$ are weight coefficients deciding which part of the cost is more important, and $T$ is the time horizon. 

It seems that the SoC balancing problem described in Eq.~\eqref{equ_10} is a simple constrained optimization problem that can be easily solved by a traditional numerical method. However, in a decentralized system, every ESU is unable to get the global information of the DESS. Thus, these constraints are hard to follow. In addition, due to the same reason, the average SoC value in the objective function is also intractable for each individual. 

\section{Decentralized Multi-agent Reinforcement Learning}\label{Dec-MARL}
In this section, the SoC balancing problem is formulated as a Markov decision process, and then Dec-MARL is developed to find the solution in a fully-decentralized way.

\subsection{Markov Decision Process Formulation for SoC Balancing Problem}\label{MDP}
The Markov decision process of the SoC balancing problem can be written as a tuple, i.e.,  $<\mathcal{G},\mathcal{S},\mathcal{O},\mathcal{A},\mathcal{R},\mathit{p}>$. In this tuple, $\mathcal{G}$ is the graph given in Sec \ref{graph theory}. 

{\it State}: The DESS consists of multiple ESUs, where each ESU operates as an RL agent that can communicate with its neighboring ESUs. Thus, the state can be defined as:
\begin{equation}
    s_t=\{E_{i,t},D_{i,t}\},\;i\in \mathcal{N}
\end{equation}
where $s_t\in \mathcal{S}$, $\mathcal{S}$ denotes the state set, and $E_{i,t}$ and $D_{i,t}$ are the SoC and the local power demand of the $i$th agent at time step $t$, respectively. 

\par {\it Observation}: Due to the decentralization of the DESS, agents are restricted to obtaining incomplete observations of the DESS state, denoted as $o_{i,t}=\{E_{i,t},D_{i,t},E_{j,t}|j\in \mathcal{N}_i\}\subseteq s_t$. 
For more precise observations, the first-order average consensus algorithm is utilized to get the average value of the global power demand and the SoCs:
\begin{equation}\label{observation}
    o_{i,t}=\{E_{i,t},D_{i,t}, E_{j,t}|j\in \mathcal{N}_i, \; \hat{E}_{i,t},\;\hat{D}_{i,t}\}
\end{equation}
where $\hat{E}_{i,t}$ and $\hat{D}_{i,t}$ denote the estimations of the average SoC and the average power demand, respectively. Note that the true state value of the DESS is observable. Hence, the observation can be considered as an element in the observation set which is a subset of the state set (i.e., $o_{i,t}\in \mathcal{O} \subset \mathcal{S}$).

\par {\it Action}: In this study, the action denotes the output power set of ESUs in the DESS:
\begin{equation}
    a_{t} = [a_{1,t},\dots ,a_{i,t},\dots,a_{N,t}]\in \mathcal{A}
\end{equation}
where $\mathcal{A}$ is the action set of the DESS, and $a_{i,t}$ is the output power of the $i$th ESU at time step $t$, i.e., $a_{i,t}=P_{i,t}$.

{\it State transition}: As the SoC of an ESU changes according to the action $a_{i,t}$ (i.e., output power $P_{i,t} $) and the local power demand given by the environment changes with time, the state transition can be given as:
\begin{equation}
    s_{t+1}\sim \mathit{p}(\cdot|a_t,s_t)
\end{equation}
where $s_t,s_{t+1} \in \mathcal{S}$ and the state transition is calculated according to (\ref{ESUM}).

{\it Reward}: After reaching the next state $s_{t+1}$, each agent obtains a local reward based on its output power and state:
\begin{equation}
    r_{i,t}=r(s_{t+1},P_{i,t})=\alpha(E_{i,t+1}-\hat{E}_{i,t})^2+\beta C^{ct}_{i,t}
\end{equation}
where $ C^{ct}_{i,t}$ can be calculated by (\ref{battery degradation1}) and (\ref{battery degradation2}). For cooperation, the agent reward is chosen as the average value of the local reward, i.e., $\overline{r}_{i,t} = \frac{1}{N}\sum_{j=1}^N r_{j,t}$.
The acquisition of the total reward $\mathcal{R}$ can be derived by considering the rewards of all agents at all time steps:
\begin{equation}
    \mathcal{R} = \sum^T_{t=1}\sum_{i=1}^N (\alpha(E_{i,t+1}-\hat{E}_{i,t})^2+\beta C^{ct}_{i,t}).
\end{equation}

\subsection{Dec-MARL}

\subsubsection{Framework}
\begin{figure}[!t]
    \centering
    \includegraphics[width=3.5in]{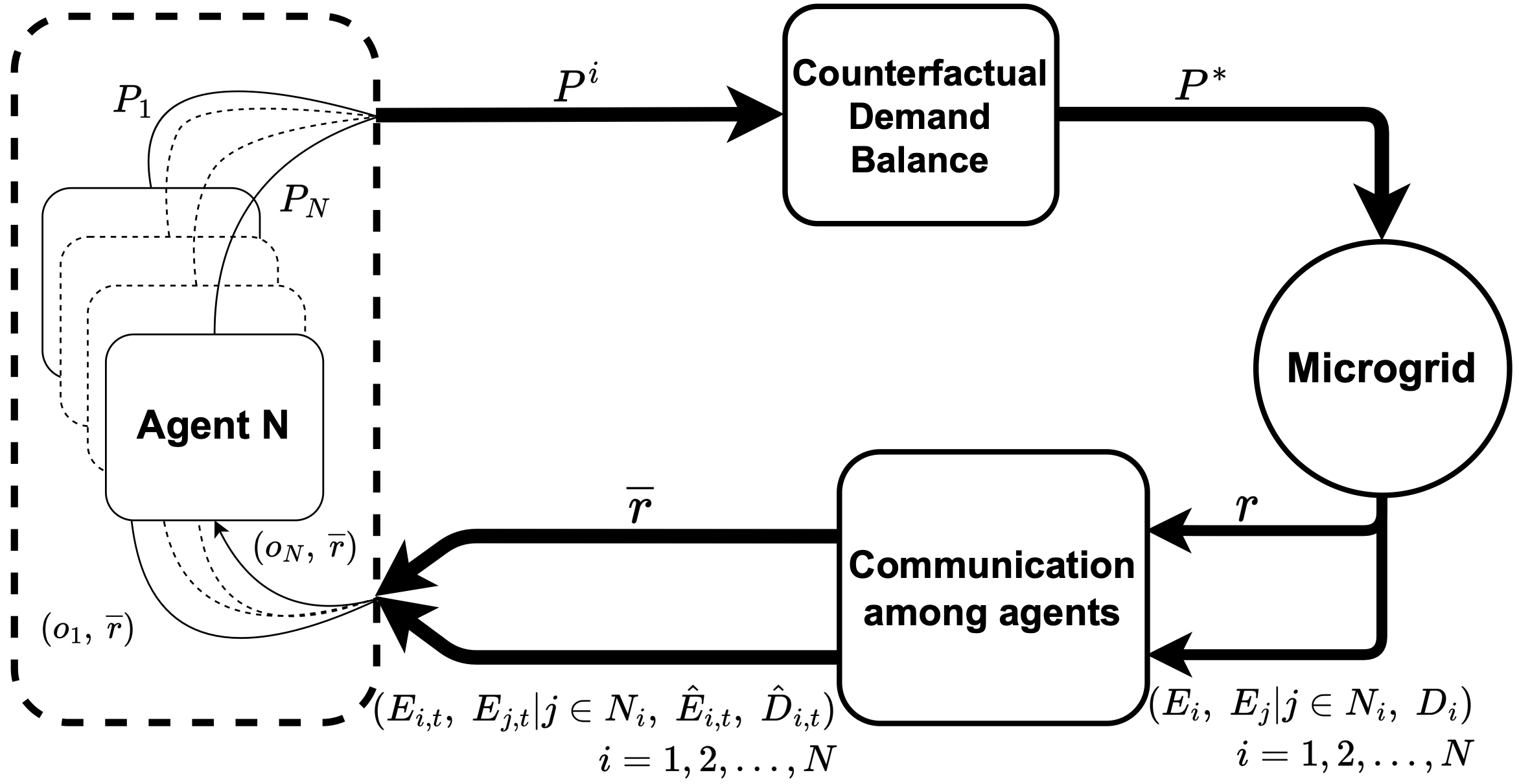}
    \caption{Framework of Dec-MARL.}
    \label{framework}
\vspace{-0.7cm}\end{figure}

The algorithm framework is described in Fig. \ref{framework}. 
Firstly, the first-order average consensus algorithm is used to get the average reward and expand the agent's observation for more effective actions through communication among agents. 
Secondly, a counterfactual demand balance (CDB) algorithm is proposed to settle the action constraints in the MARL. Finally, the Dec-MARL algorithm is proposed. Note that the DDPG is utilized for each agent in Dec-MARL. 

\subsubsection{Communication among Agents}
Based on the first-order average consensus algorithm in Sec \ref{average consensus}, it is feasible to acquire the average value of any variables within the DESS by means of communication among neighbors.
Given a weight matrix $W$ that satisfies the convergence condition, the estimation of the average power demand $\hat{D}_{i,t}$ can be obtained by the following iteration:
\begin{equation}\label{eq_3.16}
	\hat{D}_{i,t}[k+1]=\hat{D}_{i,t}[k]+\sum_{j\in \mathcal{N}_i} w_{ij}(\hat{D}_{j,t}[k]-\hat{D}_{i,t}[k]) .
\end{equation}

\par Then, $\hat{D}_{i,t}$ can be derived as follows:
\begin{equation}\label{eq_3.17}
	\hat{D}_{i,t}=\lim_{k \to \infty}\hat{D}_{i,t}[k] = \frac{1}{N}\sum_{i=1}^N D_{i,t},\quad i=1,2,\dots,N.
\end{equation}
\par Furthermore, the estimated average value of the SoCs (i.e., $\hat{E}_{i,t}$) and the reward (i.e., $\hat{r}_{i,t}$) can be obtained in the same way. Therefore, the observation can be expanded according to (\ref{observation}).

\subsubsection{Counterfactual Demand Balance Algorithm}\label{db}
Demand balance as the constraints in (\ref{equ_10}) is critical to the SoC balancing problem. 
The majority of demand balance algorithms typically adjust the output power based on the discrepancy between the power demand and the total output power, which is a challenging task in the context of the DESS with DTDE mode. 
Inspired by \cite{liu2018distributed}, the CDB algorithm is proposed to expand the exploration range while following the action constraints from the demand balance. 

In the beginning, the initial action is given by the agent's actor with the exploration noise $\mathcal{N}_t$, (i.e., $ a_{i,t}=P_{i,t}=\pi_i(o_i)+\mathcal{N}_t$), where $\pi_i(\cdot)$ is the policy of $i$th agent. 
It should be noted that the output power $P_{i,t}$ is an intermediate variable that is not executed at time step $t$. 
Next, the agent calculates the difference between the local demand and the action, that is $d_{i,t}=D_{i,t}-P_{i,t}$. 
Subsequently, the first-order average consensus algorithm is employed to compute the average value of $d_{i,t}$:
\begin{equation}\label{eq_3.18}
	\hat{d}_{i,t}= \frac{1}{N}(\sum_{i=1}^N D_{i,t}-\sum_{i=1}^NP_{i,t}).
\end{equation}
\par Once getting the estimated average difference $\hat{d}_{i,t} $, the output power can be updated as:
\begin{equation}\label{eq_3.19}
	P_{i,t}' = P_{i,t} + \text{sign}(\hat{d}_{i,t}) n_1 \max \{|\hat{d}_{i,t}|,\Delta P\}
\end{equation}
where $n_1$ is sampled from a uniform distribution, i.e., $\textbf{U}(0,1)$, and $\text{sign}(\cdot)$ is the sign function.

To ensure that the output power of ESUs remains within the allowable range, a counterfactual approach is employed:
\begin{equation}\label{eq_counterfactual drag}
P_{i,t}'=\left\{
\begin{aligned}
&n_2 \overline{P_{i,t}} & P_{i,t}'<\underline{P_{i,t}}\\
&P_{i,t}' &  P_{i,t}' \in [\underline{P_{i,t}},\overline{P_{i,t}}]\\
&n_2 \underline{P_{i,t}} & P_{i,t}'>\overline{P_{i,t}}
\end{aligned}
\right.
\end{equation}
where $n_2$ is sampled from $\textbf{U}(0,1)$. It reflects that the proposed CDB algorithm restricts the action space of RL agents to the allowable range and guides them to explore more feasible actions within that range. Specifically, when the output power is less than the lower bound, it will be dragged into the range $[0,\overline{P_{i,t}}]$, and vice versa. 
The RL agents' output power is then repeatedly adjusted according to (\refeq{eq_3.19}) and (\refeq{eq_counterfactual drag}), until the average difference between the total output power and demand is below a small threshold $\epsilon$. 
Afterward, the final output power is then acquired and denoted by $P_{i,t}^*$, which is the output power executed in the microgrid. 

The details of the proposed CDB algorithm are summarized in Algorithm \ref{demand_balance}. 
The CDB algorithm's benefits include restricting the RL agents' actions to the allowable range, exploring more feasible actions within that range, and accelerating the training speed of RL agents. It is worth noting that the efficiency of this algorithm is verified in Section \ref{Simulations}. 

\begin{algorithm}[t]
	\caption{Counterfactual Demand Balance Algorithm}
	\label{demand_balance}
	\begin{algorithmic}[1] 
    \STATE Set the minimum update value of output power $\Delta P$ and a small threshold $\epsilon$
	\FOR{agent i to N}
	\STATE Get the initial action $ a_{i,t}=P_{i,t}=\pi_i(o_i)+\mathcal{N}_t$ according to the current policy and exploration noise $\mathcal{N}_t$
	\STATE Compute the local difference between local demand and output power $d_{i,t}=D_{i,t}-P_{i,t}$
	\STATE Obtain the average value of the difference between local demand and output power $ \hat{d}_{i,t} = \frac{1}{N}(\sum_{j=1}^N d_{j,t})$
	\WHILE{$ \hat{d}_{i,t}>\epsilon$}
		\STATE Update output power according to (\ref{eq_3.19})
		\STATE Drag the updated power $P_{i,t}'$ into the allowable range through Section \ref{db} 
		\STATE Compute the new difference value $d_{i,t}'$ and then obtain the average value $\hat{d}_{i,t}' $ through communication 
		\STATE $P_{i,t} \gets P_{i,t}'$ \:,\:$ d_{i,t} \gets d_{i,t}'$
	\ENDWHILE
	\STATE $P_{i,t}^* \gets P_{i,t}$
	\ENDFOR
	\end{algorithmic}
\end{algorithm}

\begin{algorithm}[t]
\caption{Decentralized Multi-Agent Reinforcement Learning}
\label{algo:DRL}
\begin{algorithmic}
\FOR {$i=1$ to $N$}
\STATE Randomly initialize critic network $ Q_i(o_i,a_i|\theta_i^Q)$ and actor network $ \pi_i(o_i|\theta_i^{\pi})$ with weights $\theta_i^Q$ and $\theta_i^{\pi}$ for the $i$th agent\\
\STATE Initialize target network $Q_i'$ and $\pi_i'$ with weights $\theta_i^{Q'} \gets \theta_i^Q$, $\theta_i^{\pi'} \gets \theta_i^{\pi}$\\
\STATE Initialize replay buffer $R_i$ 
\ENDFOR
\STATE Initialize average consensus weight matrix through \eqref{eq_3.13} according to the predefined topological structure\\

\FOR {$episode=1$ to $E$}
\STATE Initialize state $s_0$
\FOR {$t=1$ to $T$}
\STATE For each agent $i$, obtain the local demand value $ \hat{D}_{i,t}$, neighbor SoC value $ E_{j\in \mathcal{N}_i,t}$ and average SoC value $ \hat{E}_{i,t}$ through communication
\STATE For each agent $i$, select the initial action $ a_{i,t}=P_{i,t}=\pi_i(o_{i,t}|\theta_i^{\pi})+\mathcal{N}_t$ w.r.t. the current policy
\STATE Get the final action $ a_{i,t}=P_{i,t}^*$ through Algorithm \ref{demand_balance}
\STATE Execute final actions $a_{1,t},\dots,a_{N,t}$ 
\STATE Observe local rewards $r_{1,t},\dots,r_{N,t}$ and the next state $s_{t+1}$
\STATE Obtain average reward values $\hat{r}_{1,t}, \dots,\hat{r}_{N,t}$ and the next observations $o_{1,t+1},\dots,o_{N,t+1}$ through communication
\STATE Store $ ( o_{i,t},a_{i,t},\hat{r}_{i,t},o_{i,t+1} )$ in replay buffer $R_i$
\FOR {agent $i$ to $N$}
\STATE Sample a random mini-batch of $\mathcal{M}$ samples $ ( o_i^j,a_i^j,\hat{r}_i^j,{o'_i}^j )$ from replay buffer $R$
\STATE Set $y_i^j = \hat{r}_i^j + \gamma Q_i({o'_i}^j,a'_i)|_{a'_i =\pi_i({o'_i}^j|\theta_i^{\pi}) }$
\STATE Update critic by minimizing the loss:\\
 		$\qquad\quad L(\theta_i^Q)=\frac{1}{\mathcal{M}} \sum_j (y_i^j-Q_i(o_i^j,a_i^j|\theta_i^Q))^2$
\STATE Update actor using the sampled policy gradient:\\
		$ \nabla_{\theta^{\pi}_i}J\approx \frac{1}{\mathcal{M}}\sum_j \nabla_{\theta^{\pi}_i}\pi_i(o_i^j|\theta^{\pi}_i)\nabla_{a_i^j}Q_i^{\pi_i}(o_i^j,a)|_{a = \pi_i(o_i^j)}$
\ENDFOR
\STATE Update target network parameters for $i$th agent:\\
				$\qquad\qquad\qquad {\theta^Q_i}' \gets \tau \theta^Q_i+(1-\tau){\theta^Q_i}'  $\\
				$\qquad\qquad\qquad {\theta^\pi_i}' \gets \tau \theta^\pi_i+(1-\tau){\theta^\pi_i}'  $\\
\ENDFOR
\ENDFOR
\end{algorithmic}
\end{algorithm}

\subsubsection{Decentralized Q Value Function}\label{Sec_4.4}
In Dec-MARL, the DDPG algorithm is utilized for every single agent. 
In DDPG, RL agent maintains four networks: critic network $Q_i$, critic target network $Q_i'$, actor network $ \pi_i$, actor target network $ \pi_i'$ parameterized by $ \theta_i^{Q},\theta_i^{Q'},\theta_i^{\pi},\theta_i^{\pi'}$, respectively.
Once getting the state-action transition $ ( o_i,a_i,\hat{r}_i,o'_i )$, the agent can use the value to calculate its local $Q$ target value via:
\begin{equation}
    y_i = \hat{r}_i + \gamma Q_i'(o'_i,a'_i)|_{a'_i =\pi_i(o'_i|\theta_i^{\pi'}) }
\end{equation}
where $y_i $ denotes the target value of $i$th agent, $Q_i'(\cdot )$ denotes the target network for $i$th agent, $o_i'$ denotes the observation of the next state after executing action $a_i$, and $a_i'$ is the next action chosen by agent's target policy $\pi_i'$.
\par After getting the local $Q$ target value, the parameters of critic network $\theta_i^{Q}$ can be updated by minimizing the loss function as follows:
\begin{equation}
    L(\theta_i^Q)=\frac{1}{S} \sum_S (y_i-Q_i(o_i,a_i|\theta_i^Q))^2.
\end{equation}
\par Then, the actor network can be updated by the sampled policy gradient:
\begin{equation}
    \nabla_{\theta^{\pi}_i}J\approx \frac{1}{S}\sum_S \nabla_{\theta^{\pi}_i}\pi_i(o_i|\theta^{\pi}_i)\nabla_{a_i}Q_i^{\pi_i}(o_i,a)|_{a = \pi_i(o_i|\theta^{\pi}_i)}.
\end{equation}
\par Finally, the target network will be softly updated:
\begin{equation}
    \left\{
        \begin{aligned}
            &{\theta^Q_i}' &\gets &\tau \theta^Q_i+(1-\tau){\theta^Q_i}'  \\
            &{\theta^\pi_i}' &\gets &\tau \theta^\pi_i+(1-\tau){\theta^\pi_i}'
        \end{aligned}
    \right. 
\end{equation}
where $\tau $ is the soft update rate. 
During the training process, the critic and actor networks' weights are updated by ascending in gradient. This eventually leads the $Q$ value and actor to converge at an optimal solution. 
Algorithm \ref{algo:DRL} presents the comprehensive Dec-MARL algorithm, which can be applied to derive the SoC balancing strategy for each ESU.

\begin{figure}[!t]
	\centering	\includegraphics[width=3in]{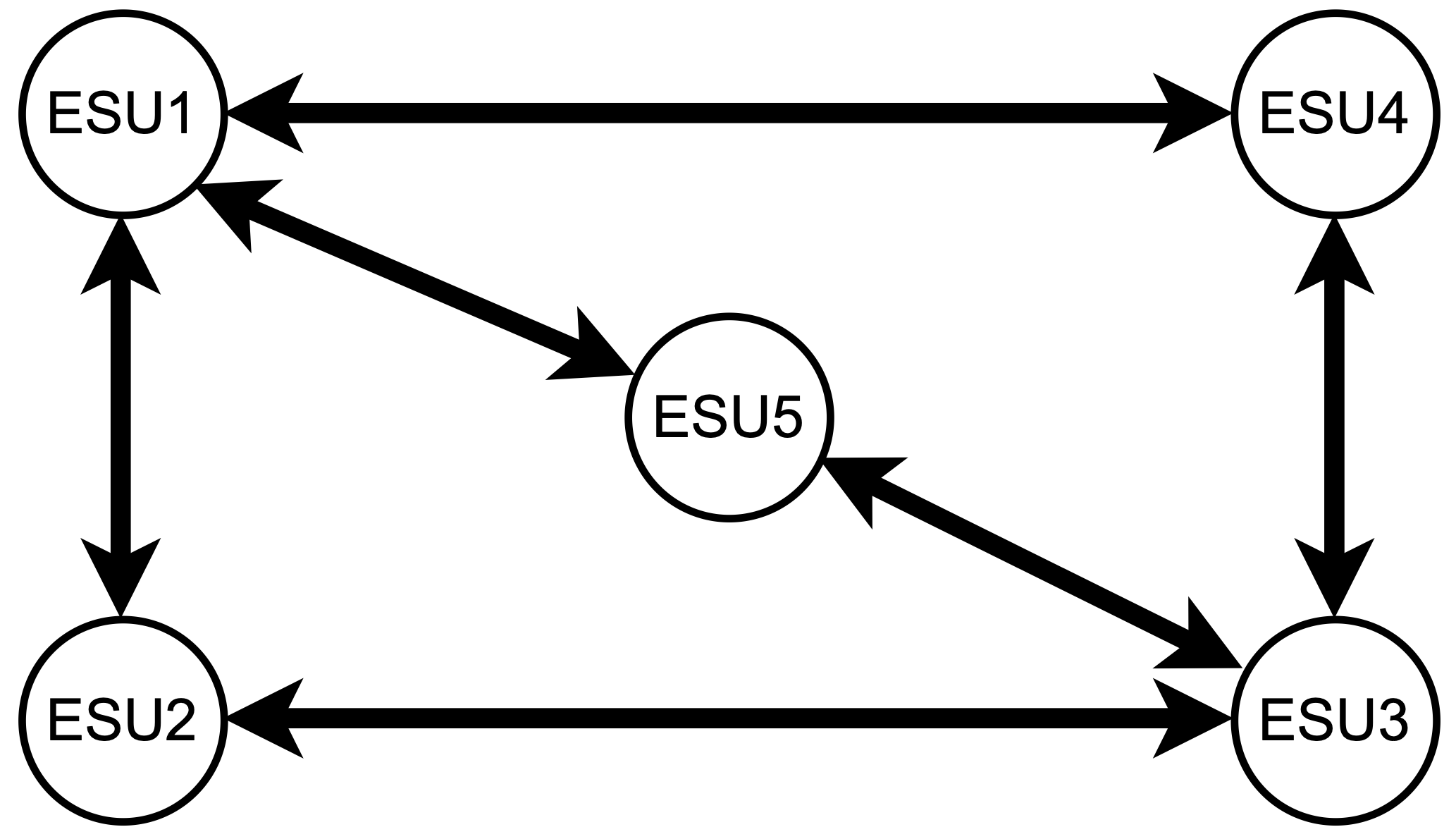}
	\caption{The DESS communication model.}\label{fig_communication5}
\vspace{-1cm}\end{figure}

\begin{figure}[t]
\centering
	\subfloat[Reward]{\includegraphics[height = 0.2\textwidth]{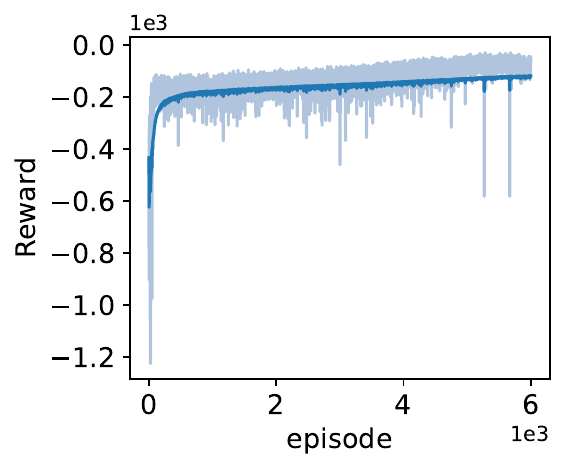}}
	\hfill
	\subfloat[SoC variance]{\includegraphics[height = 0.2\textwidth]{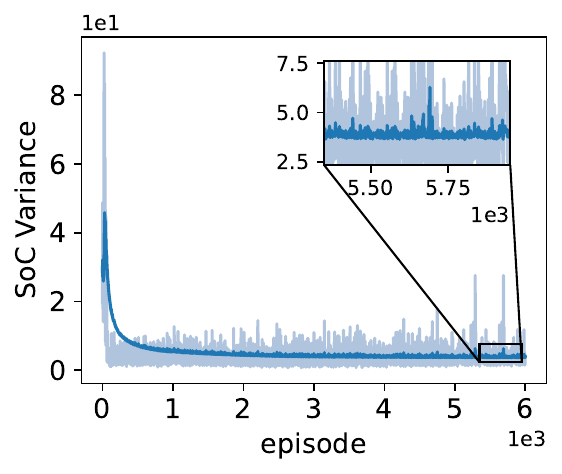}}
\caption{Training results.}
\label{training}
\vspace{-0.7cm}\end{figure}
\section{Simulation}\label{Simulations}
To investigate its effectiveness and practicality, the Dec-MARL algorithm was tested in a simple microgrid DESS model through extensive simulations. 
Meanwhile, the performance of the key component of Dec-MARL (i.e., the CDB algorithm) was demonstrated. All simulations were implemented in Python on a PC with i5-11400F CPU, 16 GB of RAM, and 64-bit operating system. 

\subsection{Training}

A DESS with five ESUs was adopted, as shown in Fig. \ref{fig_communication5}.
To satisfy the convergence condition of the first-order average consensus algorithm, the communication weight matrix $W$ needs to be designed. 
In this paper, the \textit{Metropolis-Hastings} weights proposed in \cite{xiao2007distributed} were utilized: 
\begin{equation}\label{eq_3.13}
w_{ij}=\left\{
\begin{aligned}
&1/(max\{d_i,d_j\}+1) & i\neq j, (v_i,v_j)\in \mathcal{E}\\
&1-\sum_{j\in \mathcal{N}_i}1/(max\{d_i,d_j\}+1) & i=j\\
&0 & i\neq j, (v_i,v_j)\notin \mathcal{E}
\end{aligned}
\right.
\end{equation}
where $d_i$ is the degree of $i$th ESU, which is the number of neighbors of the $i$th ESU in the DESS. 
On the basis of the communication graph given in Fig. \ref{fig_communication5}, the weight matrix $W$ was given as:
$$
W = 
\begin{bmatrix*}[r]
	0.25 & 0.25  &  0    & 0.25& 0.25\\
    0.25 &  0.5  &  0.25 &   0 & 0   \\
    0    & 0.25  &  0.25 & 0.25& 0.25\\
    0.25 &  0    & 0.25  & 0.5 & 0   \\
	0.25 & 0     &0.25   & 0   & 0.5
\end{bmatrix*}.
$$

\par In a single episode, each agent was required to make 1440 decisions within a horizon of one day, with the time step of 1 minute. Due to variations in capacities and power ranges among ESUs used in practice, parameters had been established and are presented in TABLE  \ref{tab:DESS parameters} for reference.

\begin{table}[t]
    \centering
    \caption{DESS parameters}
    \scalebox{0.95}
{    \begin{tabular}{ccccc}
        
	\toprule
	ESU&Capacity(kWh)&Power range(kW)&SoC range&charging efficiency\\
	\midrule
	ESU1&700&[-180,180]&\multirow{5}{*}{[0.1,0.9]}&\multirow{5}{*}{0.99}\\
	ESU2&1000&[-300,300]&\\
	ESU3&1200&[-360,360]&\\
	ESU4&1500&[-480,480]&\\
	ESU5&1800&[-600,600]&\\
	\bottomrule
    \end{tabular}}
    \label{tab:DESS parameters}
\end{table}
\begin{table}[t]
    \centering
    \caption{Agent learning parameters}
    \scalebox{1.25}{
    \begin{tabular}{ccccc}
		\toprule
		Learning rate& Replay buffer& $\alpha$ & $\beta$&initial noise \\
		\midrule
		0.001&30000&-200&-0.5&5\\

		\bottomrule
	\end{tabular}}

    \label{tab:learning parameters}
\end{table}

\begin{figure*}[t]
\centering
	\subfloat[Demand/output power in the performance test.]{\includegraphics[width = 0.35\textwidth]{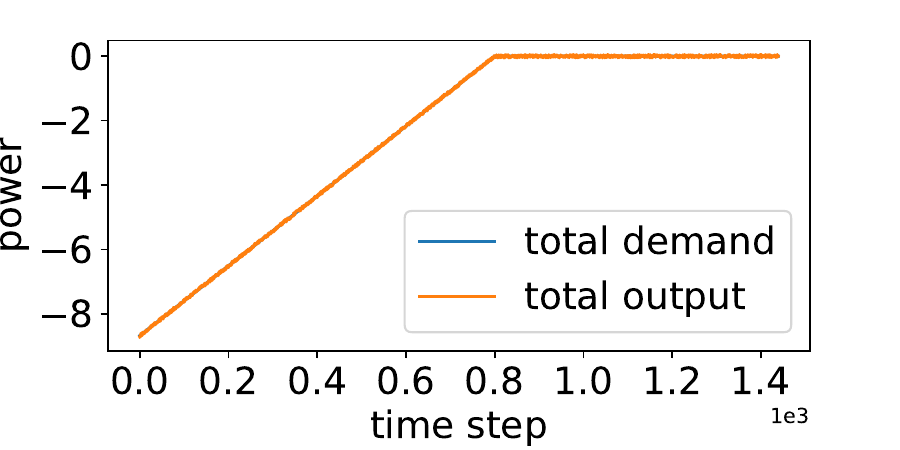}}\hspace{-10mm}
	\hfill
	\subfloat[SoC in the performance test.]{\includegraphics[width = 0.35\textwidth]{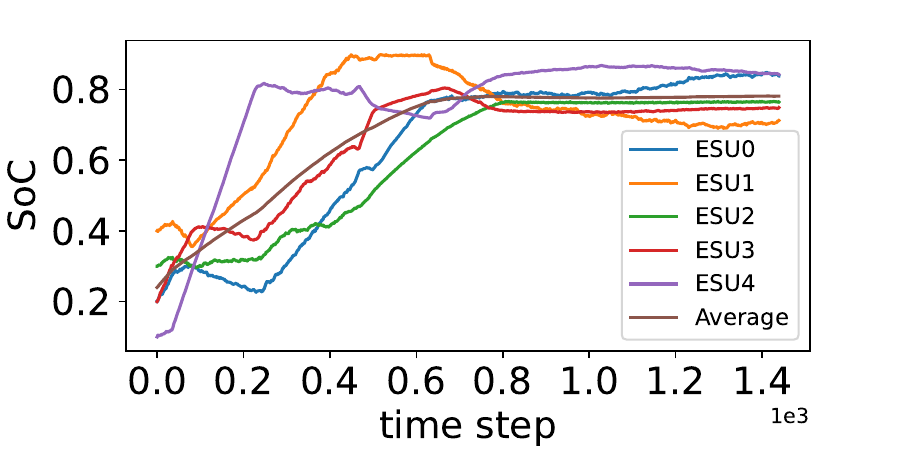}}\hspace{-10mm}
        \hfill
	\subfloat[SoC variance in the performance test.]{\includegraphics[width = 0.33\textwidth]{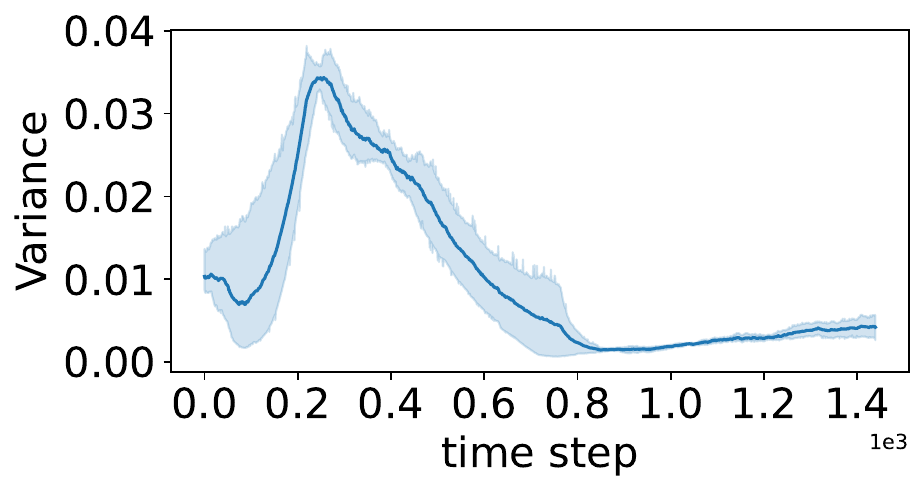}} 
\vspace{-3mm}\\
 	\subfloat[Demand/output power in the comparison test.]{\includegraphics[width = 0.34\textwidth]{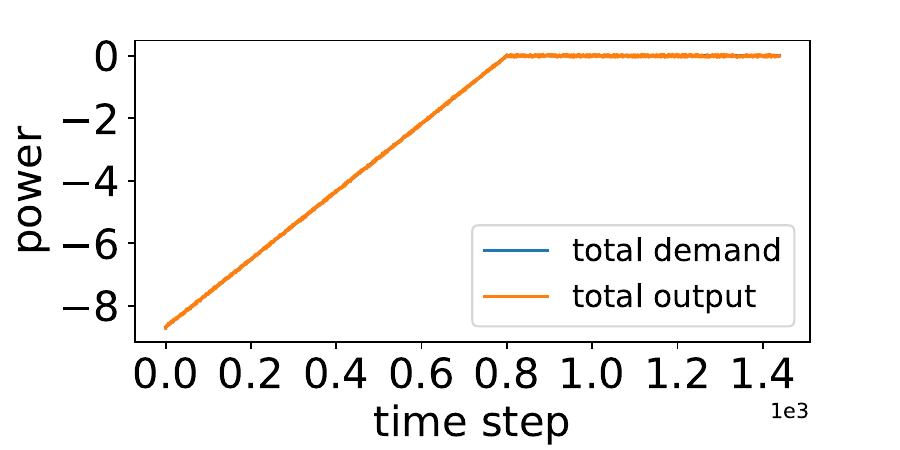}}\hspace{-10mm}
        \hfill
	\subfloat[SoC in the comparison test.]{\includegraphics[width = 0.34\textwidth]{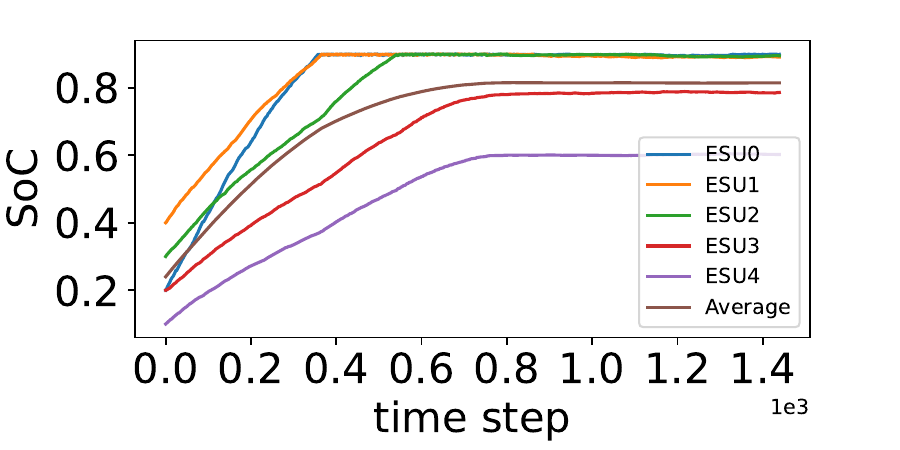}}\hspace{-10mm}
	\hfill
	\subfloat[SoC variance compare between baseline and Dec-MARL.]{\includegraphics[width = 0.35\textwidth]{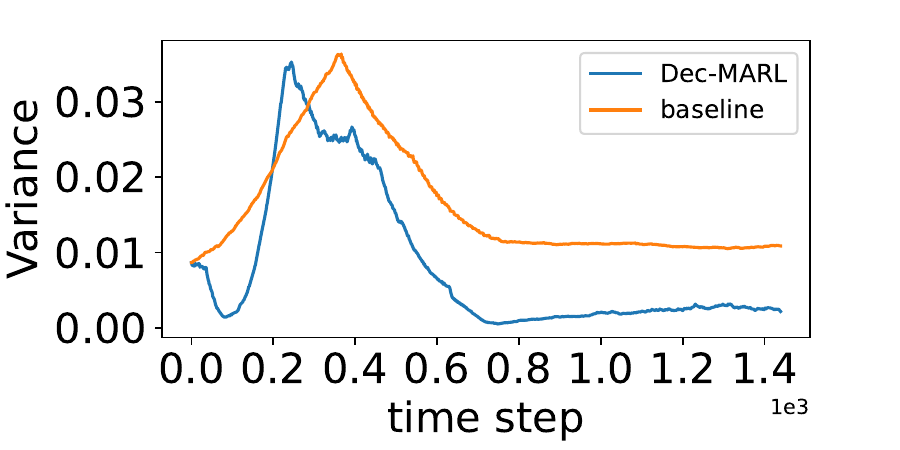}}
\caption{Simulation results in the performance test and comparison test.}
\label{fig:simulation}
\vspace{-8mm}\end{figure*}

\begin{figure}[t]
    \centering
    \includegraphics[height=0.23\textwidth]{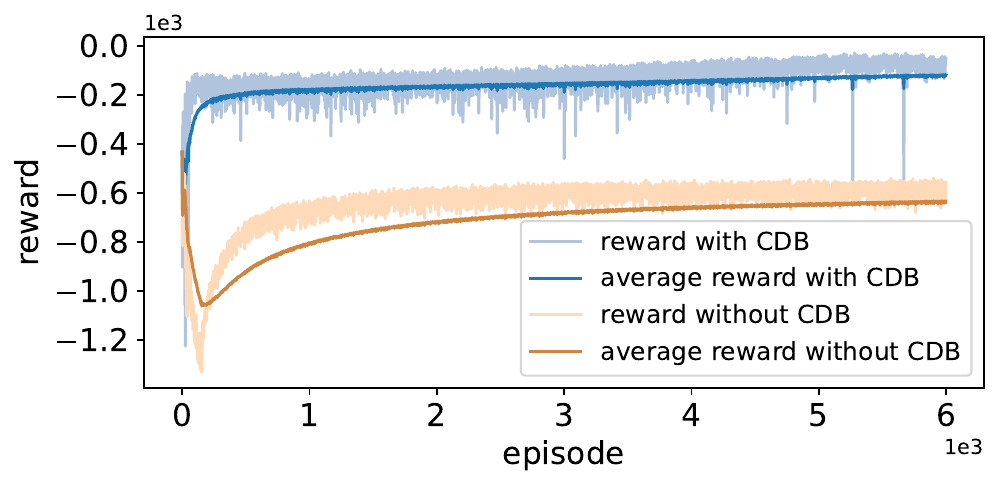}
    \caption{Performance of the CDB algorithm.}
    \label{CDB performance}
\vspace{-0.8cm}\end{figure}

The initial SoC of each ESU was set randomly in the range of $[0.7,0.9]$, and the power demand $D_T(t)\:$ (kWh) was set as:
\begin{equation}
	D_T(t) = 3\sin (\frac{t \pi}{720}).    
\end{equation}
\par Additionally, the parameters of the RL agents were designed as follows.
In this decentralized learning process, each ESU as a single agent had its own replay buffer. 
Table \ref{tab:learning parameters} displays the identical sizes of all replay buffers and learning parameters, which were designed uniformly due to the synchronized training steps and universal objective.

\par The training results are shown in Fig. \ref{training}, 
where the increase of the reward denotes a reduction in battery degradation and deviation of the SoCs.
As shown in Fig. \ref{training} (a), the total reward of the DESS converges to its maximum value after 6000 episodes of training.
Additionally, Fig. \ref{training} (b) shows a gradual reduction in the variance of the SoCs, indicating that the agents had learned to balance the SoCs while fulfilling the power demand successfully.

\subsection{Performance}
To evaluate the performance of Dec-MARL, we conducted tests on trained agents in a simplified charging simulation environment with well-defined power demand. As shown in Fig. \ref{fig:simulation} (a), the total power demand increases during the time period between $0$ and $830$ linearly, after which it remains constant at $0$ for the remaining time. The total power demand is totally covered by the total output power. Additionally, the five SoC values were initialized at random within the range $[0.1,0.5]$. The changes in the SoC are described in Fig. \ref{fig:simulation} (b) which shows the SoC values converge to a similar value (i.e., $0.8$). As the total demand was changing, each ESU chose its own action according to its observation to satisfy the total demand and tried to minimize the SoC variance. As shown in Fig. \ref{fig:simulation} (c), the SoC variance gradually decays to a steady small value. To sum up, Dec-MARL is capable of effectively balancing the SoC of DESS while meeting the power requirements. 

\subsection{Comparison between Dec-MARL and Baseline}
In the previous MARL methods, a precise action constraint is normally converted to a soft constraint, written as the penalty in the reward function. However, these methods may not remain effective for extended periods of time. This is due to training being paused when an agent's SoC exceeds allowable limits - a constraint that cannot be accommodated in real-world scenarios. In view of this, Dec-MARL was compared with a centralized proportional allocation method with the proposed CDB algorithm, which is denoted as the baseline. 

In this scenario, both methods were initialized with identical SoC values (i.e., $[0.2,0.4,0.3,0.2,0.1]$). Although the total output power of the baseline method can match the total demand during the test, it fails to address the problem of balancing the SoC of the DESS, as shown in Fig. \ref{fig:simulation} (d) and (e). Since ESUs allocated the output power proportionally, ESUs with large SoC values would increase to the upper bound of the SoC allowable range first. Without adjustment, the SoC values of these ESUs remain constant at this bound, irrespective of SoC balancing, as shown in Fig. \ref{fig:simulation} (e). For a more direct comparison, the SoC variance of Dec-MARL and the baseline method are described in Fig. \ref{fig:simulation} (f). As shown in Fig. \ref{fig:simulation} (f), the SoC variance of Dec-MARL finally converges to a lower value than the baseline method.

\subsection{Effectiveness of the CDB}
 To verify the efficiency of the CDB algorithm, a factual demand balance algorithm was defined, wherein the dragging process was set differently from the counterfactual dragging process in (\ref{eq_counterfactual drag})
\begin{equation}\label{eq_5.1}\small
	P_{i,t}[k+1]=\left\{
	\begin{aligned}
	& \underline{P_{i,t}}& P_{i,t}[k+1]<\underline{P_{i,t}}\\
	&P_{i,t}[k+1] &  P_{i,t}[k+1] \in [\underline{P_{i,t}},\overline{P_{i,t}}]\\
	&\overline{P_{i,t}} & P_{i,t}[k+1]>\overline{P_{i,t}}.
	\end{aligned}
	\right.
\end{equation}

In this factual process, the out-of-range output power would be dragged to the nearest allowable bound. It is important to note that this is the only difference between the two demand balance algorithms being compared. 
After 6000 episodes of training, the cumulative rewards were demonstrated in Fig. \ref{CDB performance}. 

As shown in Fig. \ref{CDB performance}, it's easy to find out that the DESS with the CDB algorithm performs better than the DESS with the factual demand balance algorithm. 
With the counterfactual mechanism, the agents can explore more feasible actions, and the learning speed is higher in the preliminary stage. 
Finally, the DESS of Dec-MARL converges to a higher reward compared with the competitor.

\section{Conclusions} \label{conclusion}
This paper presented a novel approach to address the SoC balancing problem in the DESS by formulating it into a finite Markov decision process with discrete time steps. A MARL method (i.e., Dec-MARL) was proposed to solve the problem in decentralized training and decentralized execution mode.
 The utilization of the first-order average consensus algorithm enabled the expansion of DESS observations, thereby facilitating the acquisition of a more precise policy. Then, a counterfactual demand balance algorithm was designed to solve action constraints in MARL by dragging the actions into the allowable range in a counterfactual way, which expands the exploration range of the RL agent and accelerates the learning speed in the preliminary stage. The proposed Dec-MARL based on average reward can be utilized to maximize the global reward while satisfying the power demand. Simulation results in various cases verified that Dec-MARL is an effective decentralized method for reducing the SoC variance among distributed ESUs in the DESS. For potential future works, the grid-connected mode of the microgrid would be considered, and energy trading can also be incorporated into the policy design.

\vspace{-0.2cm}
\renewcommand\refname{References}

\bibliographystyle{IEEEtr}
\bibliography{References}

\begin{thebibliography}{10}

\bibitem{lidula2011microgrids}
N.~Lidula and A.~Rajapakse, ``Microgrids research: A review of experimental microgrids and test systems,'' {\em Renewable and Sustainable Energy Reviews}, vol.~15, no.~1, pp.~186--202, 2011.

\bibitem{8825535}
Z.~Taylor, H.~Akhavan-Hejazi, and H.~Mohsenian-Rad, ``Optimal operation of grid-tied energy storage systems considering detailed device-level battery models,'' {\em IEEE Transactions on Industrial Informatics}, vol.~16, no.~6, pp.~3928--3941, 2020.

\bibitem{9349186}
S.~Som, S.~De, S.~Chakrabarti, S.~R. Sahoo, and A.~Ghosh, ``A robust controller for battery energy storage system of an islanded ac microgrid,'' {\em IEEE Transactions on Industrial Informatics}, vol.~18, no.~1, pp.~207--218, 2022.

\bibitem{TIE10006735}
H.~Qing, C.~Zhang, J.~Xu, S.~Zeng, and X.~Guo, ``A nonlinear multimode controller for seamless off-grid of energy storage inverter under unintentional islanding,'' {\em IEEE Transactions on Industrial Electronics}, 2023.
\newblock doi:{ \color{blue}\href{ https://doi.org/10.1109/TIE.2022.3232639 } {10.1109/TIE.2022.3232639 } }.

\bibitem{morstyn2016control}
T.~Morstyn, B.~Hredzak, and V.~G. Agelidis, ``Control strategies for microgrids with distributed energy storage systems: An overview,'' {\em IEEE Transactions on Smart Grid}, vol.~9, no.~4, pp.~3652--3666, 2016.

\bibitem{TIEcarrasco2006power}
J.~M. Carrasco, L.~G. Franquelo, J.~T. Bialasiewicz, E.~Galv{\'a}n, R.~C. PortilloGuisado, M.~M. Prats, J.~I. Le{\'o}n, and N.~Moreno-Alfonso, ``Power-electronic systems for the grid integration of renewable energy sources: A survey,'' {\em IEEE Transactions on Industrial Electronics}, vol.~53, no.~4, pp.~1002--1016, 2006.

\bibitem{ferahtia2022optimal}
S.~Ferahtia, A.~Djeroui, H.~Rezk, A.~Houari, S.~Zeghlache, and M.~Machmoum, ``Optimal control and implementation of energy management strategy for a {DC} microgrid,'' {\em Energy}, vol.~238, Article:121777,2022.

\bibitem{TIE9424403}
X.~Li, L.~Lyu, G.~Geng, Q.~Jiang, Y.~Zhao, F.~Ma, and M.~Jin, ``Power allocation strategy for battery energy storage system based on cluster switching,'' {\em IEEE Transactions on Industrial Electronics}, vol.~69, no.~4, pp.~3700--3710, 2022.

\bibitem{morstyn2017multiwithoutRL}
T.~Morstyn, A.~V. Savkin, B.~Hredzak, and V.~G. Agelidis, ``Multi-agent sliding mode control for state of charge balancing between battery energy storage systems distributed in a {DC} microgrid,'' {\em IEEE Transactions on Smart Grid}, vol.~9, no.~5, pp.~4735--4743, 2017.

\bibitem{TIESOCsingle2022yang}
F.~Yang, F.~Gao, B.~Liu, and S.~Ci, ``An adaptive control framework for dynamically reconfigurable battery systems based on deep reinforcement learning,'' {\em IEEE Transactions on Industrial Electronics}, vol.~69, no.~12, pp.~12980--12987, 2022.

\bibitem{TIESOCGuangzeShi2021ADS}
G.~Shi, H.~Han, Y.~Sun, Z.~Liu, M.~Zheng, and X.~Hou, ``A decentralized {SOC} balancing method for cascaded-type energy storage systems,'' {\em IEEE Transactions on Industrial Electronics}, vol.~68, no.~3, pp.~2321--2333, 2021.

\bibitem{jiang2019hierarchical}
W.~Jiang, C.~Yang, Z.~Liu, M.~Liang, P.~Li, and G.~Zhou, ``A hierarchical control structure for distributed energy storage system in {DC} micro-grid,'' {\em IEEE Access}, vol.~7, pp.~128787--128795, 2019.

\bibitem{TII9099954}
G.~Dong, F.~Yang, K.-L. Tsui, and C.~Zou, ``Active balancing of lithium-ion batteries using graph theory and a-star search algorithm,'' {\em IEEE Transactions on Industrial Informatics}, vol.~17, no.~4, pp.~2587--2599, 2021.

\bibitem{9720160}
H.~Wu, L.~Chai, and Y.-C. Tian, ``Distributed multirate control of battery energy storage systems for power allocation,'' {\em IEEE Transactions on Industrial Informatics}, vol.~18, no.~12, pp.~8745--8754, 2022.

\bibitem{TIEMGRL2022demandresponse}
R.~Lu, R.~Bai, Z.~Luo, J.~Jiang, M.~Sun, and H.-T. Zhang, ``Deep reinforcement learning-based demand response for smart facilities energy management,'' {\em IEEE Transactions on Industrial Electronics}, vol.~69, no.~8, pp.~8554--8565, 2022.

\bibitem{liu2018distributed}
Z.~Liu, Y.~Luo, R.~Zhuo, and X.~Jin, ``Distributed reinforcement learning to coordinate current sharing and voltage restoration for islanded {DC} microgrid,'' {\em Journal of Modern Power Systems and Clean Energy}, vol.~6, no.~2, pp.~364--374, 2018.

\bibitem{sutton2018reinforcement}
R.~S. Sutton and A.~G. Barto, {\em Reinforcement learning: An introduction}.
\newblock MIT press, 2018.

\bibitem{single2019intelligent}
Y.~Du and F.~Li, ``Intelligent multi-microgrid energy management based on deep neural network and model-free reinforcement learning,'' {\em IEEE Transactions on Smart Grid}, vol.~11, no.~2, pp.~1066--1076, 2019.

\bibitem{single2020deep}
F.~S. Gorostiza and F.~M. Gonzalez-Longatt, ``Deep reinforcement learning-based controller for {SoC} management of multi-electrical energy storage system,'' {\em IEEE Transactions on Smart Grid}, vol.~11, no.~6, pp.~5039--5050, 2020.

\bibitem{TIEsingle2022HEV}
B.~Hu and J.~Li, ``A deployment-efficient energy management strategy for connected hybrid electric vehicle based on offline reinforcement learning,'' {\em IEEE Transactions on Industrial Electronics}, vol.~69, no.~9, pp.~9644--9654, 2022.

\bibitem{multi2020SamadiCTCE}
E.~Samadi, A.~Badri, and R.~Ebrahimpour, ``Decentralized multi-agent based energy management of microgrid using reinforcement learning,'' {\em International Journal of Electrical Power \& Energy Systems}, vol.~122, Article:106211,2020.

\bibitem{li2019multiEM}
F.~Li, J.~Qin, and W.~X. Zheng, ``Distributed {Q}-learning-based online optimization algorithm for unit commitment and dispatch in smart grid,'' {\em IEEE Transactions on Cybernetics}, vol.~50, no.~9, pp.~4146--4156, 2019.

\bibitem{zhu2022multiEM}
D.~Zhu, B.~Yang, Y.~Liu, Z.~Wang, K.~Ma, and X.~Guan, ``Energy management based on multi-agent deep reinforcement learning for a multi-energy industrial park,'' {\em Applied Energy}, vol.~311, Article:118636,2022.

\bibitem{CTDE2021multiEM}
X.~Fang, Q.~Zhao, J.~Wang, Y.~Han, and Y.~Li, ``Multi-agent deep reinforcement learning for distributed energy management and strategy optimization of microgrid market,'' {\em Sustainable Cities and Society}, vol.~74, Article:103163,2021.

\bibitem{xuCTDE2020multiHEM}
X.~Xu, Y.~Jia, Y.~Xu, Z.~Xu, S.~Chai, and C.~S. Lai, ``A multi-agent reinforcement learning-based data-driven method for home energy management,'' {\em IEEE Transactions on Smart Grid}, vol.~11, no.~4, pp.~3201--3211, 2020.

\bibitem{single2018HEV}
R.~Xiong, J.~Cao, and Q.~Yu, ``Reinforcement learning-based real-time power management for hybrid energy storage system in the plug-in hybrid electric vehicle,'' {\em Applied Energy}, vol.~211, pp.~538--548, 2018.

\bibitem{xiao2007distributed}
L.~Xiao, S.~Boyd, and S.-J. Kim, ``Distributed average consensus with least-mean-square deviation,'' {\em Journal of Parallel and Distributed Computing}, vol.~67, no.~1, pp.~33--46, 2007.

\bibitem{antoniadou2020market}
K.~Antoniadou-Plytaria, D.~Steen, O.~Carlson, M.~A.~F. Ghazvini, {\em et~al.}, ``Market-based energy management model of a building microgrid considering battery degradation,'' {\em IEEE Transactions on Smart Grid}, vol.~12, no.~2, pp.~1794--1804, 2020.

\end{thebibliography}

\end{document}